# Improving spam filtering by combining Naive Bayes with simple k-nearest neighbor searches

Daniel Etzold
etzold@cs.tum.edu

September 25, 2018

**Abstract**
Using naive Bayes for email classification has become very popular within the last few months. They are quite easy to implement and very efficient. In this paper we want to present empirical results of email classification using a combination of naive Bayes and k-nearest neighbor searches. Using this technique we show that the accuracy of a Bayes filter can be improved slightly for a high number of features and significantly for a small number of features.

## 1   Introduction

Proposed by Sahami et al [3] in the year 1998, naive Bayes [5, 6] have been used very successfully for spam filtering within the last few months. Although filter systems based on naive Bayes achieve a very high overall accuracy of about 97% [4], in this paper we want to try to improve these results by combining naive Bayes with simple k-nearest neighbor searches. As shown in [2] the accuracy of naive Bayes can be increased successively by increasing the dimension of the document vectors until the dimension (number of features) reaches about 1600. We now show that with the approach presented in this paper the dimension of document vectors can be reduced down to 500 features. At the same time we achieve an even higher classification accuracy than without the combination of naive Bayes and a k-nearest neighbor search.

The k-nearest neighbor search (kNN) [7, 8] is also a very simple method to classify documents and was found to show very good performance on text categorization tasks. It is a "lazy-learning" method, which means that it does not need a learning phase. The only thing which has to be done is to index the documents of the training set and convert them into a document vector representation. When classifying a new document the similarity of its document vector to each document vector in the training set has to be computed. Then, we determine the categories of the $k$ nearest neighbors and choose the category which occurs most frequently. As a measure for the similarity of two documents we use the widely used Euclidean distance and the angle between two documents.

The paper is organized as follows. In the next section we describe our simple approach to combine naive Bayes with a k-nearest neighbor search. In the third section we describe our experiments and the way we tokenize emails and in the last section we present the results of our experiments.

## 2    Combining naive Bayes with kNN

Let $Pr_{nb}[\mathcal{M} = S]$ and $Pr_{nb}[\mathcal{M} = G]$ be the probability computed by using naive Bayes that an email $\mathcal{M}$ is spam / legitimate. Of course, it follows that $Pr_{nb}[\mathcal{M} = S] + Pr_{nb}[\mathcal{M} = G] = 1$ for the case of spam filtering.
Furthermore, let $\vec{u}, \vec{v} \in \mathbb{R}^V$ be two document vectors and $V$ the number of features or dimension of the vectors. The Euclidean distance of $\vec{u}$ and $\vec{v}$ is computed as follows:

$$e(\vec{u}, \vec{v}) := ||\vec{u} - \vec{v}|| = \sqrt{(u_1 - v_1)^2 + \ldots + (u_V - v_V)^2}$$

The probabilities $Pr_{knn}[\mathcal{M} = S]$ and $Pr_{knn}[\mathcal{M} = G]$ that an email $\mathcal{M}$ is classified as spam / legitimate using the k-nearest neighbor classification are computed as follows. First, we are looking for the $k$ nearest neighbors of $\vec{\mathcal{M}}$. Here, $\vec{\mathcal{M}}$ denotes the document vector of an email $\mathcal{M}$. Then, we determine the categories of the emails in the neighborhood. Let $n_S$ be the number of emails assigned to the category "spam" and $n_G$ be the number of emails assigned to the category "legitimate".

Now, the probabilities $Pr_{knn}[\mathcal{M} = S]$ and $Pr_{knn}[\mathcal{M} = G]$ can be computed as follows:

$$Pr_{knn}[\mathcal{M} = S] = \frac{n_S}{k}, \qquad Pr_{knn}[\mathcal{M} = G] = \frac{n_G}{k} = 1 - \frac{n_S}{k}$$

The next step of the algorithm is to combine the probabilities of the naive Bayes classification and the k-nearest neighbor search. We compute a score $\delta_G$ for the category "legitimate" and $\delta_S$ for the category "spam" as follows:

$$\delta_G = \frac{\alpha Pr_{nb}[\mathcal{M} = G] + \beta Pr_{knn}[\mathcal{M} = G]}{\alpha + \beta}, \qquad \delta_S = \frac{\alpha Pr_{nb}[\mathcal{M} = S] + \beta Pr_{knn}[\mathcal{M} = S]}{\alpha + \beta}$$

**Theorem:** $\delta_G + \delta_S = 1$

**Proof:**

$$\frac{1}{\alpha+\beta}\left(\alpha Pr_{nb}[\mathcal{M}=G]+\beta Pr_{knn}[\mathcal{M}=G]+\alpha Pr_{nb}[\mathcal{M}=S]+\beta Pr_{knn}[\mathcal{M}=S]\right)=$$

$$\frac{1}{\alpha+\beta}\left(\alpha\underbrace{(Pr_{nb}[\mathcal{M}=G]+Pr_{nb}[\mathcal{M}=S])}_{=1}+\beta\underbrace{(Pr_{knn}[\mathcal{M}=G]+Pr_{knn}[\mathcal{M}=S])}_{=1}\right)=$$

$$\frac{\alpha+\beta}{\alpha+\beta}=1$$

In our experiments we choose $\alpha=\beta=1$ and compute the score $\delta_G$ for every new document. We assign the document to the category "legitimate" if $\delta_G \geq 0.5$, otherwise to the category "spam".

### 2.1 Using angles instead of Euclidean distances

In this paper we do not only use the Euclidean distance as a measure of similarity between two documents but we also perform $k$-nearest neighbor searches by computing the angle between two documents.

The angles between two vectors $\vec{u}$ and $\vec{v}$ can be computing by the following equation:

$$\mathbf{sim}(\vec{u},\vec{v}) := \triangleleft(\vec{u},\vec{v}) = \frac{\vec{u}\bullet\vec{v}}{||\vec{u}||\cdot||\vec{v}||} = \frac{\sum_i u_i v_i}{\sqrt{\sum_i u_i^2}\cdot\sqrt{\sum_i v_i^2}} \quad (1)$$

## 3 Experiments

The email corpus consists of 15950 emails. 5000 of these emails are assigned to the category "spam", the remaining 10950 emails are assigned to the category "legitimate". We run our experiments with three different training and test sets. The first pair of training and test set is created by splitting the corpus at a ratio of 25:75. The second pair is at a ratio of 30:70 and the third one at a ratio of 40:60.

For each pair we reduce the dimension by using information gain to 500, 1000, 1500 and 2000 features. Thus, we get 12 models which we test with different values for parameter $k$ of the k-nearest neighbor search.

To tokenize each email we take the header and body, convert them to lower case, remove all html tags (all characters between '<...>') and extract all words w([a-z]{2,}) and numbers ([0-9]{2,}) with at least two characters. We ignore uuencoded lines.

Since spam is often written in html, we append the word "html" three times at the end of the email for each occurring html tag which occurs. In previous experiments we have achieved good results with this technique.

"legitimate" → "legitimate"

| k | V = 500 | | | V = 1000 | | | V = 1500 | | | V = 2000 | | |
|---|---|---|---|---|---|---|---|---|---|---|---|---|
|   | 0.25 | 0.3 | 0.4 | 0.25 | 0.3 | 0.4 | 0.25 | 0.3 | 0.4 | 0.25 | 0.3 | 0.4 |
| - | .9795 | .9774 | .9781 | .9849 | **.9840** | .9843 | .9867 | .9654 | .9858 | .9865 | .9859 | .9851 |
| 1 | **.9910** | **.9905** | **.9912** | **.9922** | .9725 | .9878 | **.9918** | **.9918** | **.9921** | .9898 | **.9915** | **.9923** |
| 2 | .9888 | .9881 | .9889 | .9898 | .9740 | .9871 | .9895 | .9903 | .9910 | **.9906** | .9902 | .9910 |
| 3 | .9875 | .9867 | .9874 | .9895 | .9760 | .9874 | .9899 | .9892 | .9896 | **.9906** | .9896 | .9896 |
| 4 | .9861 | .9858 | .9868 | .9890 | .9765 | .9875 | .9896 | .9892 | .9892 | .9904 | .9893 | .9893 |
| 5 | .9850 | .9841 | .9842 | .9881 | .9768 | .9875 | .9889 | .9884 | .9883 | **.9906** | .9884 | .9880 |

"spam" → "spam"

| k | V = 500 | | | V = 1000 | | | V = 1500 | | | V = 2000 | | |
|---|---|---|---|---|---|---|---|---|---|---|---|---|
|   | 0.25 | 0.3 | 0.4 | 0.25 | 0.3 | 0.4 | 0.25 | 0.3 | 0.4 | 0.25 | 0.3 | 0.4 |
| - | .9525 | .9560 | .9550 | .9648 | .9666 | **.9657** | .9680 | .9706 | .9677 | **.9698** | .9766 | .9710 |
| 1 | .9632 | .9654 | .9663 | .9632 | .9654 | .8540 | .9611 | .9649 | .9600 | .8746 | .9691 | .9650 |
| 2 | **.9680** | **.9918** | **.9720** | **.9715** | .9749 | .8957 | **.9731** | .9746 | .9717 | .8963 | **.9786** | .9747 |
| 3 | .9667 | .9648 | .9710 | **.9715** | **.9769** | .8967 | .9715 | **.9749** | .9730 | .9003 | .9783 | **.9757** |
| 4 | .9667 | .9691 | .9680 | .9712 | .9763 | .8977 | .9717 | .9743 | **.9737** | .9040 | .9783 | .9750 |
| 5 | .9669 | .9691 | .9660 | .9712 | .9760 | .9000 | .9709 | .9746 | .9727 | .9131 | .9780 | .9750 |

Figure 1: These tables show the classification accuracy which is achieved by combining naive Bayes with a kNN search which uses the Euclidean distance as the similarity measure. The table at the top shows the accuracy that legitimate emails were classified correctly. The table at the bottom shows the accuracy that spam was classified correctly.

To classify an email of the test set with naive Bayes we use rainbow [1], a toolkit to perform statistical text classification. With the options "–lex-white" and "–lex-pipe-command" rainbow can be told to use our tokenizer instead of the build-in tokenizer.

To compute the k-nearest neighbors of an email we first convert all emails of the training set into their document vector representation and compute the Euclidean distance or the angle between each of these vectors and the document vector of the email which we want to classify. For simplicity, the attributes of the vectors contain the number of occurrences of the represented feature. We do not apply and special weighting algorithm.

We combine the results of the naive Bayes and k-nearest neighbor classification as described in the previous section and choose the category with the highest $\delta$ score.

## 4 Results

The results of our experiments are shown in figure 1 for which we have used the Euclidean distance as the measure of similarity and figure 2 for which we have used the angle between two document vectors as the measure of similarity.
In both figures the table at the top shows the accuracy that legitimate emails were classified correctly, whereas the accuracy that spam was classified correctly is shown in the table at the bottom. The maximum of each column is bold. The first row which

"legitimate" → "legitimate"

| k | V = 500 | | | V = 1000 | | | V = 1500 | | | V = 2000 | | |
|---|---|---|---|---|---|---|---|---|---|---|---|---|
| | 0.25 | 0.3 | 0.4 | 0.25 | 0.3 | 0.4 | 0.25 | 0.3 | 0.4 | 0.25 | 0.3 | 0.4 |
| - | .9795 | .9774 | .9781 | .9849 | .9841 | .9843 | .9867 | .9860 | .9858 | .9865 | .9859 | .9851 |
| 1 | **.9910** | **.9906** | **.9910** | **.9921** | **.9914** | **.9913** | **.9920** | **.9918** | **.9920** | **.9921** | **.9915** | **.9925** |
| 2 | .9888 | .9881 | .9887 | .9898 | .9902 | .9904 | .9898 | .9905 | .9909 | .9903 | .9902 | .9912 |
| 3 | .9875 | .9867 | .9877 | .9895 | .9894 | .9891 | .9900 | .9892 | .9895 | .9896 | .9896 | .9900 |
| 4 | .9860 | .9858 | .9869 | .9889 | .9888 | .9889 | .9898 | .9892 | .9890 | .9898 | .9894 | .9895 |
| 5 | .9850 | .9686 | .9842 | .9879 | .9875 | .9875 | .9889 | .9884 | .9883 | .9888 | .9885 | .9881 |

"spam" → "spam"

| k | V = 500 | | | V = 1000 | | | V = 1500 | | | V = 2000 | | |
|---|---|---|---|---|---|---|---|---|---|---|---|---|
| | 0.25 | 0.3 | 0.4 | 0.25 | 0.3 | 0.4 | 0.25 | 0.3 | 0.4 | 0.25 | 0.3 | 0.4 |
| - | .9525 | .9560 | .9550 | .9648 | .9666 | .9657 | .9680 | .9706 | .9677 | .9699 | .9766 | .9710 |
| 1 | .9632 | .9651 | .9657 | .9640 | .9617 | .9597 | .9611 | .9646 | .9620 | .9621 | .9674 | .9647 |
| 2 | **.9664** | **.9703** | **.9723** | .9712 | .9717 | .9707 | **.9731** | .9746 | .9723 | **.9739** | **.9786** | .9750 |
| 3 | .9651 | .9697 | .9710 | **.9717** | .9720 | **.9723** | .9717 | **.9751** | .9737 | .9731 | .9783 | **.9760** |
| 4 | .9632 | .9680 | .9697 | **.9717** | **.9723** | **.9723** | .9717 | .9749 | **.9747** | .9720 | .9783 | .9753 |
| 5 | .9632 | .9686 | .9677 | **.9717** | **.9723** | .9720 | .9709 | **.9751** | **.9747** | .9715 | .9780 | .9753 |

Figure 2: These tables show the classification accuracy which is achieved by combining naive Bayes with a kNN search which uses the angle between two documents as the similarity measure. The table at the top shows the accuracy that legitimate emails were classified correctly. The table at the bottom shows the accuracy that spam was classified correctly.

is labeled as "-" contains the results of a naive Bayes classification without a k-nearest neighbor search.

In most cases (21 of 24) the combination of naive Bayes with a k-nearest neighbor search performs better than a classification without using a k-nearest neighbor search. Both similarity measures (Euclidean distance and the angle between documents) achieve similar results.

In practice it is quite important to prevent wrong classification of legitimate emails because this is much more expensive than the wrong classification of spam. A legitimate email which is incorrectly assigned to the category "spam" is called *false positive*. Our results show that the number of *false positives* can be reduced dramatically by our new classification approach. For the case of correctly classified legitimate emails we achieve in most experiments a classification accuracy better than 99% for $k = 1$.

Also for $k = 1$ and $V = 500$ the classification accuracy of legitimate emails is higher than the accuracy for $V = 2000$ without a k-nearest neighbor search. Therefore, the dimension or number of features can be reduced down to 500 which results in lower computational cost for the learning and classification task.

Nevertheless, for both similarity measures the best classification accuracy of legitimate emails is still achieved for a large number of features, here for $V = 2000$ and $k = 1$. The accuracy that is achieved is 99.23% for the Euclidean distance and 99.25% by using the angle as similarity measure.

In contrast to legitimate emails for which the best accuracy is obtained for $k = 1$, the best accuracy for spam (for a small number of features) is obtained for $k = 2$. As previously seen for legitimate emails, also for spam this accuracy is higher than the accuracy which is achieved for the classification without our new approach for a high number of features ($V = 2000$).

Therefore, when using our approach to reduce the computational cost for learning and classification it should be the best to use only 500 features with a value for $k$ either equal to one or to two. The number of false positives is reduced most for a value of one, whereas for a value of two, both, the accuracy for legitimate emails and spam, is better than the accuracy of a classification without our approach for high dimensional document vectors.